\documentclass[final,12pt]{elsarticle}

\makeatletter
\def\ps@pprintTitle{%
  \let\@oddhead\@empty
  \let\@evenhead\@empty
  \def\@oddfoot{Published in Information Sciences (https://doi.org/10.1016/j.ins.2014.11.046)\hfil}
  \let\@evenfoot\@oddfoot
}
\makeatother

\usepackage{graphicx}

\usepackage{amssymb}

\usepackage[caption=false,font=footnotesize]{subfig}
\usepackage[ruled,noline]{algorithm2e}

\newcommand{\bbbr}{\mathbb{R}}

\journal{Information Sciences}

\begin{document}

\begin{frontmatter}



\title{Vesicoureteral Reflux Detection with Reliable Probabilistic Outputs}

\author[fuc]{Harris Papadopoulos}
\address[fuc]{Frederick University, 7 Y. Frederickou St., Palouriotisa, Nicosia 1036, Cyprus}
\ead{h.papadopoulos@frederick.ac.cy}

\author[duth]{George Anastassopoulos}
\address[duth]{Medical Informatics Laboratory, Democritus University of Thrace, GR-68100, Alexandroupolis, Greece}
\ead{anasta@med.duth.gr}

\begin{abstract}
Vesicoureteral Reflux (VUR) is a pediatric disorder in which 
urine flows backwards from the bladder to the upper urinary tract. 
Its detection is of great importance as it increases the 
risk of a Urinary Tract Infection, which can then lead to a kidney infection since 
bacteria may have direct access to the kidneys. Unfortunately the detection of VUR
requires a rather painful medical examination, called voiding cysteourethrogram (VCUG), 
that exposes the child to radiation. In an effort to avoid the exposure to radiation 
required by VCUG some recent studies examined the use of machine learning techniques 
for the detection of VUR based on data that can be obtained without exposing the child 
to radiation. This work takes one step further by proposing an approach that provides 
lower and upper bounds for the conditional probability of a given child having VUR. 
The important property of these bounds is that they are guaranteed (up to statistical 
fluctuations) to contain well-calibrated probabilities with the 
only requirement that observations are independent and identically distributed (i.i.d.). 
Therefore they are much more informative and reliable than the plain yes/no answers 
provided by other techniques.
\end{abstract}


\end{frontmatter}


\section{Introduction}\label{sec:intro}

Vesicoureteral Reflux (VUR) is a pediatric disorder that has potentially very serious consequences as it might lead to a kidney infection (pyelonephritis). Specifically, in VUR urine flows abnormally from the bladder back into one or both ureters and, in some cases, into one or both kidneys. The severity of VUR is classified into five grades, grade I being the least severe and grade V being the most severe. The disorder increases the risk of Urinary Tract Infections (UTIs), which, if left untreated, can lead to kidney damage. Therefore young children diagnosed with UTI should be further examined for VUR. However, the principal medical examination for the detection of VUR, the \emph{voiding cysteourethrogram} (VCUG), is not only a painful procedure, but also demands the exposure of the child to radiation. For this reason, the development of a technique that would help avoid VCUG and consequently the exposure to radiation is very desirable. 

The use of machine learning techniques towards this goal using as inputs clinical and laboratorial information that can be obtained without the need of radiation exposure was examined in some recent studies \cite{drube:vur,mantza:vurnn,mantza:vuricann,mantza:vurwseas}. The techniques proposed in these studies however only provide a yes or no output, without giving any further information about how much one can rely on this output. This is of course a disadvantage of most existing medical decision support systems, as this is what most conventional machine learning techniques provide. Nevertheless it is a significant disadvantage, especially in a medical setting where some indication about the likelihood of each diagnosis is of paramount importance \cite{holst:lack}.

In this work we address this drawback with the use of a recently developed machine learning framework called \emph{Venn Prediction} (VP). Venn Prediction was proposed in \cite{vovk:venn} while a detailed description of the framework can be found in~\cite{vovk:alrw}. It provides a way of extending conventional classifiers to develop techniques that produce \emph{multiprobability predictions} without assuming anything more than i.i.d.\ observations. In effect multiprobability predictions are a set of probability distributions for the true classification of the new example, which can be summarized by lower and upper bounds for the conditional probability of each new example belonging to each one of the possible classes for the task in question. The resulting bounds are guaranteed to contain well-calibrated probabilities (up to statistical fluctuations).

Until now the VP framework has been combined with the \emph{k}-Nearest Neighbour classifier in \cite{vovk:alrw} and \cite{dash:vennit}, with Support Vector Machines in \cite{lambrou:ivpaiai,zhou:vennplatts}, with Logistic Regression in \cite{ilia:lrvpaiai} and with Artificial Neural Networks (ANN) in \cite{papa:nnvpaiai,papa:nnvp}. In this work we apply the Artificial Neural Network Venn Predictor (ANN-VP) to the problem of detecting VUR based on a dataset consisting of children diagnosed with UTI and further examined with VCUG. The data were collected by the Pediatric Clinical Information System of Alexandroupolis University Hospital, Greece. We follow a slightly modified version of the approach proposed in \cite{papa:nnvpaiai}, so as to address the class imbalance problem of the particular dataset. In particular we incorporate minority oversampling and majority undersampling in the ANN-VP and compare the performance of the two approaches. Our experimental results show that the probabilistic outputs of the ANN-VPs outperform the ones of conventional ANN in all cases, while minority oversampling performs better than majority undersampling. Furthermore we demonstrate that the probability bounds produced by Venn Prediction are well-calibrated as opposed to the ones produced by conventional ANNs which can be very misleading.

The rest of this paper starts with a review of related work on the prediction of VUR in the next section. This is followed by a description of the dataset used in this study in Section~\ref{sec:VUR}. 
Then in Section~\ref{sec:Venn} it gives overview of the Venn Prediction framework, while in Section~\ref{sec:NNVP} it 
details the proposed ANN-VP with Minority Oversampling and ANN-VP Majority Undersampling algorithms.
Section~\ref{sec:Res} presents the experiments performed in both the batch and on-line settings and reports the obtained 
results. Finally, Section~\ref{sec:Conc} gives the conclusions and future directions of this work.

\section{Related Work}

There are some studies in the literature for the prediction of VUR without the use of Artificial Intelligence techniques, which however seem to have very low specificity. The two most recent ones are \cite{leroy:vur,sorkhi:vur}. In \cite{leroy:vur} the authors propose a clinical decision rule for predicting VUR of grade III or higher. The rule was derived using a total of 494 patients and it had a sensitivity of $86\%$ and a specificity of $43\%$ in internal cross-validation. In \cite{sorkhi:vur} the authors explore the use of Dimercaptosuccinic Acid scan and Ultrasonography for predicting VUR. They report a sensitivity of $81\%$ and a specificity of $53\%$.

The only study we were able to find other than the ones of the Democritus University of Thrace group on the prediction of VUR using Machine Learning techniques is \cite{drube:vur}. In this study the authors identified a urinary proteome pattern for detecting High-grade VUR (grade IV or V) with the use of support vector machines (SVM) on data obtained by capillary electrophoresis coupled to mass spectrometry. The resulting proteome test was validated on 36 patients with $88\%$ sensitivity and $79\%$ specificity.

The studies \cite{mantza:vurnn,mantza:vuricann,mantza:vurwseas} were performed on the same dataset we use here, however only a particular part of it consisting of $20\%$ of the cases was used for testing. In \cite{mantza:vurwseas} the authors studied the use of multilayer perceptrons, in \cite{mantza:vuricann} the authors studied the use of probabilistic neural networks (PNNs) and in \cite{mantza:vurnn} the PNNs were combined with a genetic algorithm for optimization of the feature subset and parameters used. The best results were achieved in \cite{mantza:vurnn}, with a best accuracy of $96.3\%$ on the particular test set. However when tested on the whole dataset in a 10-fold cross-validation setting all PNNs proposed had very low sensitivity.

It's worth to note that none of the approaches proposed in these studies provide any type of probabilistic outputs which is the aim of this work. 

\section{Vesicoureteral Reflux Disease Data}\label{sec:VUR}

\begin{table}[t]
  \centering
  \caption{VUR clinical and laboratorial parameters together with their values}
  \label{tab:data}
  \resizebox{\textwidth}{!}{%
  \begin{tabular}{@{\extracolsep{0.15cm}}l l l l l l l l l} \hline\noalign{\smallskip}
  No. & Parameter & \multicolumn{7}{l}{Possible values} \\ 
  \noalign{\smallskip}\hline\noalign{\smallskip}
  1     & Sex        & Boy       & Girl            & & & & & \\
  2     & Age        & $<\!1$ year & 1-5 years     & $>\!5$ years  & & & & \\
  3     & Siblings   & $1$       & $2$             & $3$         & & & & \\
  4-8   & Systsymp   & Fever     & \multicolumn{2}{c}{Vomit/diarrhea}  & Anorexia    & \multicolumn{2}{c}{Weight loss} & Others \\
  9     & WBC        & $<\!4500$   & \multicolumn{2}{c}{4500-10500}    & \multicolumn{1}{l}{$>\!10500$}    &   &   & \\
  10    & WBC type   & n         & L               & m           & E & b & & \\
  11    & Ht         & $<\!37$     & 37-42           & $>\!42$       &   &   & & \\
  12    & Hb         & $<\!11.5$   & 11.5-13.5       & $>\!13.5$     &   &   & & \\
  13    & PLT        & $<\!170$    & 170-450         & $>\!450$      &   &   & & \\
  14    & ESR        & $<\!20$     & 20-40           & $>\!40$       &   &   & & \\
  15    & CRP        & $+$       & $-$             &             &   &   & & \\
  16    & Bacteria   & E.coli    & Proteus         & Kiebsielas  & Strep       & Stapf & Psedom & Other \\
  17-22 & Sensitiv   & Penicillin & Kefalosp2      & Kefalosp3   & Aminoglyc   & \multicolumn{2}{c}{Sulfonamides} & Other \\
  23    & Ultrasound & Rsize nrm  & Rsize abn      & Rstract nrm & Rstract abn & Normal         & Other & \\
  24    & Dursymp    & $2$ days   & $3$ days       & $4$ days    & $5$ days    & $>\!5$ days      & & \\
  25    & Starttre   & $2$ days   & $3$ days       & $4$ days    & $5$ days    &                & & \\
  26-27 & Riskfact   & \multicolumn{2}{l}{Age $<\!1$ year} & Ttreat    &             &             &               & \\
  28    & Collect    & U-bag      & Catheter       & Suprapubic  &             &                & & \\
  29-34 & Resistan   & Penicillin & Kefalosp2      & Kefalosp3   & Aminoglyc   & \multicolumn{2}{c}{Sulfonamides}  & Other \\
  \noalign{\smallskip}\hline\noalign{\smallskip}
  \end{tabular}}
\end{table}

\begin{table}[t]
  \centering
  \caption{Features selected by the three feature selection techniques}
  \label{tab:featsel}
  \begin{tabular}{@{\extracolsep{0.15cm}}l l} \hline\noalign{\smallskip}
  \,Technique & Selected features \\ \noalign{\smallskip}\hline\noalign{\smallskip}
  \,CFS & 11, 21, 22, 23 and 27 \\
  \,$\chi^2$/IG & 11, 12, 15, 19, 23, 27 and 28 \\ \noalign{\smallskip}\hline
  \end{tabular}
\end{table}

The VUR data used in this study, were obtained from the Pediatric Clinical Information System of Alexandroupolis University Hospital, Greece. The dataset consists of 162 child patients with UTI, of which 30 were diagnosed with VUR. The clinical and laboratorial parameters considered are the ones collected according to the medical protocol of the hospital. In total there were 19 parameters, which include
\begin{itemize}
  \item general information: sex, age and number of siblings
  \item the clinical presentation (Systsymp)
  \item blood laboratory testing results: white blood cell count (WBC), WBC type, haematocrit (Ht), haemoglobine (Hb), platelets (PLT), erythrocyte sedimentation rate (ESR) and C-reactive protein (CRP)
  \item urine cultures: bacteria 
  \item antibiogramme: sensitivity and resistance to antibiotics
  \item renal / bladder ultrasound results
  \item duration of the symptoms (Dursymp)
  \item start of the treatment (Starttre)
  \item risk factors: whether the child is less than one year old and an assessment of risk by the attending clinicians (Ttreat)
  \item method of urine collection (Collect)
\end{itemize}
A list of these parameters and their values is given in Table~\ref{tab:data}. It is emphasized that some of the parameters may take more than one values simultaneously. For example the clinical presentation (Systsymp) can be a combination of symptoms. These parameters were transformed to a binary set of sub-parameters, one for each of their possible values indicating existence (1) or not (0) of the particular value. Therefore the clinical presentation was converted to five parameters, the sensitivity and resistance were converted to six parameters each and the risk factors to two parameters. For this reason these parameters have a range in the first column of Table~\ref{tab:data}. After this conversion the total number of parameters was 34.

Due to the large number of parameters and the relatively small number of cases, feature selection was applied to the data so as to avoid overfitting. Specifically, three feature selection techniques were used: correlation-based feature subset selection (CFS) \cite{Hall1998} in conjunction with best-first search and the chi-squared ($\chi^2$) and information gain (IG) feature evaluation techniques retaining the features with values above zero. The last two returned the same feature subset so the two subsets reported in table~\ref{tab:featsel} were used in our experiments.

\section{The Venn Prediction Framework}\label{sec:Venn}

This section gives a brief description of the Venn prediction framework; for
more details the interested reader is referred to \cite{vovk:alrw}. We are 
given a training set $\{(x_1, y_1), \dots, (x_l, y_l)\}$ of examples, where 
each $x_i \in \bbbr^d$ is the vector of attributes for example $i$ and 
$y_i \in \{Y_1, \dots, Y_c\}$ is the classification of that example. We are 
also given a new unclassified example $x_{l+1}$ and our task is to predict 
the probability of this new example belonging to each class 
$Y_j \in \{Y_1, \dots, Y_c\}$ based only on the assumption that 
all $(x_i, y_i), i= 1, 2, \dots$ are i.i.d.

The Venn Prediction framework assigns each one of the possible classifications 
$Y_j \in \{Y_1, \dots, Y_c\}$ to $x_{l+1}$ in turn and generates the extended set
\begin{equation}
\label{eq:extset}
  \{(x_1, y_1), \dots, (x_l, y_l), (x_{l+1}, Y_j)\}.
\end{equation}
For each resulting set (\ref{eq:extset}) it then divides the examples 
into a number of categories and calculates the probability of $x_{l+1}$ belonging 
to each class $Y_k$ as the frequency of $Y_k$'s in the category that contains it.

To divide each set (\ref{eq:extset}) into categories it uses what is called 
a \emph{Venn taxonomy}. A Venn taxonomy is a measurable function that assigns a 
category $\kappa_i^{Y_j}$ to each example $z_i$ in (\ref{eq:extset}); the output of 
this function should not depend on the order of the examples.
Every Venn taxonomy defines a different Venn Predictor. Typically each taxonomy 
is based on a traditional
machine learning algorithm, called the \emph{underlying algorithm} of the 
VP. The output of this algorithm for each attribute vector 
$x_i, i = 1,\dots, l+1$ after being trained either on the whole set (\ref{eq:extset}), 
or on the set resulting after removing the pair $(x_i, y_i)$ from (\ref{eq:extset}),
is used to assign $\kappa_i^{Y_j}$ to $(x_i, y_i)$.
For example, a Venn taxonomy that can be used with every traditional algorithm 
assigns the same category to all examples that are given the 
same classification by the underlying algorithm. The Venn taxonomy used in
this work is defined in the next section.

After assigning the category $\kappa_i^{Y_j}$ to each example $(x_i, y_i)$ in the extended set 
(\ref{eq:extset}), the empirical probability of each classification $Y_k$ among the 
examples assigned $\kappa_{l+1}^{Y_j}$ will be
\begin{equation}
\label{eq:prob}
  p^{Y_j}(Y_k) = \frac{\left|\{i = 1, \dots, l+1| \kappa_i^{Y_j} = \kappa_{l+1}^{Y_j} \,\,\&\,\, y_i = Y_k\}\right|}{\left|\{i = 1, \dots, l+1| \kappa_i^{Y_j} = \kappa_{l+1}^{Y_j}\}\right|}.
\end{equation}
This is a probability distribution for the label of $x_{l+1}$. After 
assigning all possible labels to $x_{l+1}$ we get a set of probability
distributions that compose the multiprobability prediction of the 
Venn predictor $P_{l+1} = \{p^{Y_j} : Y_j \in \{Y_1, \dots, Y_c\}\}$.
As proved in \cite{vovk:alrw} the predictions produced by any Venn predictor are 
automatically valid multiprobability predictions. This is true regardless 
of the taxonomy of the Venn predictor. Of course the taxonomy used is 
still very important as it determines how efficient, or informative, the 
resulting predictions are. We want the diameter of multiprobability 
predictions and therefore their uncertainty to be small and we also 
want predictions to be as close as possible to zero or one.

The maximum and minimum probabilities obtained for each class $Y_k$ 
define the interval for the probability of the new example belonging 
to $Y_k$:
\begin{equation}
\label{eq:interval}
\bigg[\min_{k = 1,\dots,c} p^{Y_j}(Y_k), \max_{k = 1,\dots,c} p^{Y_j}(Y_k)\bigg].
\end{equation}
If the lower bound of this interval is denoted as $L(Y_k)$ and the upper bound is 
denoted as $U(Y_k)$, the Venn predictor finds 
\begin{equation}
\label{eq:pred}
k_{best} = \arg\max_{k = 1,\dots, c} \overline{p(Y_k)},
\end{equation}
where $\overline{p(Y_k)}$ is the mean of the probabilities obtained for $Y_k$, 
and outputs the class $\hat y = Y_{k_{best}}$ as its prediction together with the 
interval $[L(\hat y), U(\hat y)]$
as the probability interval that this prediction is correct. The complementary 
interval $[1 - U(\hat y), 1 - L(\hat y)]$
gives the probability that $\hat y$ is not the true classification of 
the new example and it is called the \emph{error probability interval}.

In this case however, due to the class imbalance of the data and the much higher 
prior probability of the negative class, a threshold was used for determining the 
prediction of our Venn predictors. Specifically, a Venn predictor outputs $\hat y = 1$ 
as its prediction if the mean probability of the positive class $\overline{p(1)}$ is 
above a given threshold $\theta$ and $\hat y = 0$ otherwise. The probability interval
for the prediction being correct remains $[L(\hat y), U(\hat y)]$.

\section{Artificial Neural Networks Venn Prediction with \\Minority Oversampling or Majority Undersampling}\label{sec:NNVP}

This section describes the Venn Taxonomy used in this work, which is based on ANN, 
and gives the complete algorithm of the proposed 
approach. The ANNs used were 2-layer fully connected 
feed-forward networks with tangent sigmoid hidden units and a single logistic 
sigmoid output unit. They were trained with the variable learning rate backpropagation 
algorithm minimizing cross-entropy error. 
As a result their outputs can be interpreted as probabilities for class $1$ and 
they can be compared with those produced by the Venn predictor. Early stopping was used 
based on a validation set consisting of $20\%$ of the corresponding training set.

In order to address the class imbalance problem of the data two approaches were examined: 
minority oversampling (MO) and majority undersampling (MU). Specifically, in the case of the 
MO approach before training the ANN examples belonging to the minority class were randomly selected 
with replacement and copied again into the training set until the number of positive and negative 
examples became equal. In the case of the MU approach a random subset of the majority class 
examples equal to the size of the minority class was selected to be included in the training set 
and the remaining majority class examples were not used for training. In both cases this resulted 
in training the ANN with an equal number of positive and negative examples. 

After adding the new example $(x_{l+1}, Y_j)$ to the oversampled or undersampled training set and 
training the ANN, the output $o_i$ produced by the ANN for each input pattern $x_i$ 
was used to determine its category $\kappa_i$. Specifically the range of the ANN output 
$[0, 1]$ was split to a number of equally sized regions $\lambda$ and the same 
category was assigned to the examples with output falling in the same region.
In other words each one of these $\lambda$ regions defined one category of the taxonomy.

\begin{algorithm}
\KwIn{training set $T_o = \{(x_1, y_1), \dots, (x_l, y_l)\}$, test example $x_{new}$, number of categories $\lambda$, threshold $\theta$.}
Oversample the minority class generating the training set $T_m = \{(x_1, y_1), \dots, (x_n, y_n)\}$, where $n>l$ OR undersample the majority class generating the training set $T_m = \{(x_1, y_1), \dots, (x_n, y_n)\}$, where $n<l$\;
\For{$k = 0$ \KwTo $1$}
{
   Train the ANN on the extended $T_m$ set $\{(x_1, y_1), \dots, (x_n, y_n), (x_{new}, k)\}$\;
   Supply the input patterns $x_1, \dots, x_l, x_{new}$ to the trained ANN to obtain the outputs $o_1, \dots, o_l, o_{new}$\;
   \For{$i = 1$ \KwTo $l$}
   {
      Assign $\kappa_i$ to $(x_i, y_i)$ based on the output $o_i$\;
   }
   Assign $\kappa_{new}$ to $(x_{new}, k)$ based on the output $o_{new}$\;
   $p^{k}(1) := \frac{\left|\{i = 1, \dots, l, new| \kappa_i^{k} = \kappa_{new}^{k} \,\,\&\,\, y_i = 1\}\right|}{\left|\{i = 1, \dots, l, new | \kappa_i^{k} = \kappa_{new}^{k}\}\right|}$\;
   $p^{k}(0) := 1 - p^{k}(1)$\;
}
\eIf{$\overline{p(1)} > \theta$}{$\hat y = 1$}{$\hat y = 0$}
\KwOut\\
\Indp
Prediction $\hat y$\;
The probability interval for $\hat y$: $[\min_{k = 0, 1} p^{k}(\hat y), \max_{k = 0, 1} p^{k}(\hat y)]$.
\caption{ANN-VP with MO/MU\label{al:nnvp}}
\end{algorithm}

Using this taxonomy we assigned a category $\kappa_i^0$ to each example $(x_i, y_i)$ in the 
original training set extended with $(x_{l+1}, 0)$ and a category $\kappa_i^1$ to each 
example $(x_i, y_i)$ in the original training set extended with $(x_{l+1}, 1)$. 
It is important to note that the MO and MU approaches are only part of the taxonomy used and 
the original (non oversampled or undersampled) training set should be used for calculating the 
multiprobability predictions with~(\ref{eq:prob}).
We then followed the process described in Section~\ref{sec:Venn} to calculate the outputs of 
the ANN Venn Predictor with Minority Oversampling (ANN-VP with MO) and of the ANN Venn Predictor 
with Majority Undersampling (ANN-VP with MU), which is presented in Algorithm~\ref{al:nnvp}.

\section{Experimental Results}\label{sec:Res}

We performed experiments in both the batch setting and the on-line setting. The former is the standard 
setting for evaluating machine learning algorithms in which the algorithm is trained on a given training set 
and its performance is assessed on a set of test cases. In the on-line setting examples are predicted one by 
one and immediately after prediction their true classification is revealed and they are added to the training 
set for predicting the next example; the algorithm is re-trained each time on the growing training set. 
We use this setting to demonstrate the empirical validity of the multiprobability outputs produced by the ANN-VPs.

Before each 
training session all attributes were normalised setting their mean value to $0$ and their standard deviation to $1$. 
For all experiments with conventional ANN, minority oversampling or majority undersampling was performed on the 
training set in the same way as for the proposed approaches.

\subsection{Batch Setting Results}

This subsection examines the performance of the proposed approaches with the two feature subsets reported in 
table~\ref{tab:featsel} in the batch setting and compares their performance to that of the original ANN-VP, 
to that of the corresponding conventional ANNs and to that of the 
best approach developed in previous studies on the same data. Specifically we compare it to the third PNN proposed 
in \cite{mantza:vurnn} (we actually tried out all five PNNs of the same study and this was the one that gave the 
best results). Furthermore, it examines the effect that different choices for the number of taxonomy categories 
$\lambda$ have.

For these experiments we followed a 10-fold cross-validation process for $10$ times with 
different divisions of the dataset into the $10$ folds and the results reported here 
are the mean values over all runs. The ANNs used consisted of 5 hidden units, which is a sensible choice 
for the small number of features selected. The threshold $\theta$ used for determining the prediction for all 
three Venn predictors was set to $0.1852$, which is the frequency of the positive examples in the 
dataset.

In order to be able to use standard metrics for the evaluation of probabilistic outputs, which evaluate 
single probabilities rather than probability intervals, we convert the output of the ANN-VPs to 
$\overline{p(1)}$; corresponding to the estimate of the ANN-VP for the probability of  
each test example being positive.
For reporting these results five quality metrics are used. The first two are the sensitivity and 
specificity of each classifier, which do not take into account the probabilistic outputs produced, 
but are typical metrics for assessing the quality of classifiers. The third 
is cross-entropy error (or log-loss):
\begin{equation}
CE = -\sum^{N}_{i=1} y_i\log(\hat p_i) + (1 - y_i)\log(1 - \hat p_i),
\end{equation}
where $N$ is the number of test examples and $\hat p_i$ is the probability 
produced by the algorithm for each test example being positive; this is the error minimized by 
the training algorithm of the ANNs on the training set.
The fourth metric is the Brier score~\cite{brier:brscore}:
\begin{equation}
BS = \frac{1}{N} \sum^{N}_{i=1} (\hat p_i - y_i)^2.
\end{equation}
The cross-entropy error, or log-loss, and the Brier score are the 
most popular quality metrics for probability assessments.

The Brier score can be decomposed into three terms interpreted as 
the uncertainty, reliability and resolution of the probabilities, 
by dividing the range of probability values into a number of 
intervals $K$ and representing each interval $k = 1,\dots, K$ by a 
`typical' probability value $r_k$ \cite{murphy:brierpartition}.
The reliability term of this decomposition measures how close the output probabilities are 
to the true probabilities and therefore reflects how well-calibrated 
the output probabilities are. Since the reliability of probabilities is of 
paramount importance in a medical setting, this is the 
fifth metric used here. It is defined in \cite{murphy:brierpartition} as:
\begin{equation}
REL = \frac{1}{N} \sum^{K}_{k=1} n_k (r_k - \phi_k)^2,
\end{equation}
where $n_k$ is the number of examples with output probability in 
the interval $k$ and $\phi_k$ is the percentage of these examples 
that are positive. Here the number of categories $K$ was 
set to $20$.

\begin{table}
  \centering
  \begin{tabular}{@{\extracolsep{0.08cm}}llccccc} \hline\noalign{\smallskip}
  Features    & Algorithm   & Sens.   & Spec.   & CE     & BS     & REL \\ \noalign{\smallskip}\hline\noalign{\smallskip}
              & ANN-VP      & 64.33\% & 84.17\% & 618.45 & 0.1165 & 0.0019 \\
              & ANN-VP + MO & 75.67\% & 84.09\% & 558.92 & 0.0988 & 0.0041 \\
  \multicolumn{1}{l}{CFS}         & ANN + MO    & 70.33\% & 85.83\% & 697.67 & 0.1309 & 0.0350 \\
              & ANN-VP + MU & 72.33\% & 79.17\% & 609.57 & 0.1130 & 0.0039 \\
              & ANN + MU    & 61.33\% & 83.03\% & 921.19 & 0.1750 & 0.0569 \\ \noalign{\smallskip}\hline\noalign{\smallskip}
              & ANN-VP      & 67.00\% & 81.67\% & 614.39 & 0.1143 & 0.0027 \\
              & ANN-VP + MO & 71.33\% & 84.70\% & 595.96 & 0.1087 & 0.0052 \\
  \multicolumn{1}{l}{$\chi^2/IG$} & ANN + MO    & 66.00\% & 86.06\% & 773.30 & 0.1411 & 0.0370 \\
              & ANN-VP + MU & 66.00\% & 76.44\% & 652.33 & 0.1234 & 0.0032 \\
              & ANN + MU    & 58.00\% & 79.17\% & 897.42 & 0.1806 & 0.0576 \\ \noalign{\smallskip}\hline\noalign{\smallskip}
  \multicolumn{2}{l}{PNN 3 from \cite{mantza:vurnn}} & 53.00\% & 92.95\% & - & - & - \\ \noalign{\smallskip}\hline
  \end{tabular}
  \caption{Performance of the two ANN-VPs in the batch setting with the two different feature subsets and comparison with that of the original ANN-VP, of the corresponding conventional ANNs and of the best previously proposed approach.}
  \label{tab:allres}
\end{table}

Table~\ref{tab:allres} reports the performance of the two ANN-VPs proposed in this work together with that of the 
original ANN-VP (without MO or MU, but with the same threshold), of the corresponding conventional ANNs and of the 
best previously proposed approach for the same data. The first five rows report the results obtained using the 
features selected with the Correlation-based Feature Selection (CFS) technique, the sixth to the tenth row report the 
results obtained using the features selected based on the $\chi^2$ and Information Gain (IG) feature evaluation (since 
both methods selected exactly the same features), the last row reports the results obtained with PNN 3 from 
\cite{mantza:vurnn}, which was the best performing approach on the particular data proposed so far. 
The results of the ANN-VPs 
reported here were obtained with the number of taxonomy categories $\lambda$ set to 6, which was the value used in previous
studies \cite{papa:nnvpaiai,papa:vureann}. However as it is shown in the tables that follow the particular value does not
affect the results by much. It should be noted that for the first two metrics higher values correspond to better 
performance, whereas for the last three metrics lower values correspond to better performance.

\begin{table}[t]
  \centering
  \begin{tabular}{@{\extracolsep{0.1cm}}lrccccc} \hline\noalign{\smallskip}
  Features    & \multicolumn{1}{c}{$\lambda$} & Sens.   & Spec.   & CE     & BS     & REL \\ \noalign{\smallskip}\hline\noalign{\smallskip}
       &  2 & 77.00\% & 84.09\% & 585.40 & 0.1097 & 0.0107 \\
       &  3 & 73.33\% & 84.77\% & 564.71 & 0.1025 & 0.0047 \\
       &  4 & 75.33\% & 84.62\% & 564.82 & 0.1010 & 0.0058 \\
       &  5 & 74.67\% & 84.85\% & 563.13 & 0.1005 & 0.0055 \\
   \multicolumn{1}{l}{CFS} &  6 & 75.67\% & 84.09\% & 558.92 & 0.0988 & 0.0041 \\
       &  7 & 76.00\% & 84.55\% & 557.52 & 0.0987 & 0.0050 \\
       &  8 & 74.00\% & 84.24\% & 556.25 & 0.0979 & 0.0031 \\
       &  9 & 76.00\% & 83.56\% & 556.21 & 0.0984 & 0.0030 \\
       & 10 & 76.33\% & 84.24\% & 557.91 & 0.0991 & 0.0029 \\ \noalign{\smallskip}\hline\noalign{\smallskip}
       &  2 & 73.33\% & 83.03\% & 616.83 & 0.1155 & 0.0065 \\
       &  3 & 71.00\% & 85.61\% & 593.93 & 0.1090 & 0.0055 \\
       &  4 & 71.33\% & 84.39\% & 606.36 & 0.1109 & 0.0066 \\
       &  5 & 72.00\% & 84.55\% & 606.33 & 0.1111 & 0.0071 \\
   \multicolumn{1}{l}{$\chi^2/IG$} &  6 & 71.33\% & 84.70\% & 595.96 & 0.1087 & 0.0052 \\
       &  7 & 71.67\% & 83.79\% & 599.44 & 0.1084 & 0.0046 \\
       &  8 & 71.67\% & 83.18\% & 596.57 & 0.1080 & 0.0042 \\
       &  9 & 71.67\% & 83.26\% & 599.08 & 0.1090 & 0.0040 \\
       & 10 & 74.00\% & 83.48\% & 601.01 & 0.1090 & 0.0057 \\ \noalign{\smallskip}\hline
  \end{tabular}
  \caption{Performance of the ANN-VP with MO in the batch setting with different $\lambda$.}
  \label{tab:molambda}
\end{table}

\begin{table}[t]
  \centering
  \begin{tabular}{@{\extracolsep{0.1cm}}lrccccc} \hline\noalign{\smallskip}
  Features    & \multicolumn{1}{c}{$\lambda$} & Sens.   & Spec.   & CE     & BS     & REL \\ \noalign{\smallskip}\hline\noalign{\smallskip}
&  2 & 73.33\% & 76.97\% & 648.90 & 0.1230 & 0.0049 \\
&  3 & 66.33\% & 80.91\% & 638.80 & 0.1202 & 0.0052 \\
&  4 & 73.00\% & 79.09\% & 619.42 & 0.1155 & 0.0029 \\
&  5 & 72.67\% & 80.38\% & 615.46 & 0.1148 & 0.0046 \\
\multicolumn{1}{l}{CFS} &  6 & 72.33\% & 79.17\% & 609.57 & 0.1130 & 0.0039 \\
&  7 & 75.00\% & 80.76\% & 586.90 & 0.1078 & 0.0044 \\
&  8 & 74.33\% & 79.92\% & 597.02 & 0.1098 & 0.0036 \\
&  9 & 75.00\% & 80.38\% & 589.38 & 0.1079 & 0.0031 \\
& 10 & 75.33\% & 79.92\% & 591.32 & 0.1090 & 0.0029 \\ \noalign{\smallskip}\hline\noalign{\smallskip}
&  2 & 64.33\% & 72.65\% & 696.15 & 0.1338 & 0.0018 \\
&  3 & 64.33\% & 77.65\% & 668.30 & 0.1279 & 0.0028 \\
&  4 & 67.67\% & 77.58\% & 661.95 & 0.1258 & 0.0026 \\
&  5 & 65.67\% & 76.14\% & 659.12 & 0.1246 & 0.0034 \\
\multicolumn{1}{l}{$\chi^2/IG$} &  6 & 66.00\% & 76.44\% & 652.33 & 0.1234 & 0.0032 \\
&  7 & 65.67\% & 76.36\% & 656.83 & 0.1234 & 0.0035 \\
&  8 & 67.33\% & 75.30\% & 658.08 & 0.1235 & 0.0039 \\
&  9 & 69.00\% & 75.91\% & 659.91 & 0.1240 & 0.0034 \\
& 10 & 66.33\% & 75.68\% & 665.58 & 0.1246 & 0.0045 \\ \noalign{\smallskip}\hline
  \end{tabular}
  \caption{Performance of the ANN-VP with MU in the batch setting with different $\lambda$.}
  \label{tab:mulambda}
\end{table}

Comparing the sensitivity and specificity values reported in table~\ref{tab:allres} we see that the proposed ANN-VP 
with MO had higher sensitivity than the original ANN-VP with both feature subsets. The ANN-VP with MU on the other hand, 
had higher sensitivity than the original ANN-VP when using the feature subset selected by CFS, but a lower one when using 
the subset selected by $\chi^2$ and IG. Both proposed ANN-VPs (with MO and MU) always had higher sensitivity than the 
corresponding conventional ANNs, whereas the opposite happens in the case of specificity. In the case of the 
previously proposed PNN it seems that its sensitivity is extremely low. However, since in this work we are mainly interested 
in probabilistic outputs, the main comparison is on the last three metrics. Here the superiority of the VPs is very clear
since in all cases the VPs give lower values than the corresponding conventional ANNs. In fact the difference in reliability 
is impressive, which shows that even after reducing the probabilistic bounds produced by the two VPs to single probabilities, 
they are still very reliable.

By comparing the ANN-VP with MO to the ANN-VP with MU and the original ANN-VP approaches we see that the 
former always performs better at the CE and BS metrics. Reliability is more or less at the same levels for the three methods; 
the small differences are not really significant for this metric. The majority undersampling approach does not result 
in a clear improvement as expected. This is most likely due to the small size of the minority class, which makes the 
training set after majority undersampling very small. It would be interesting to check how its performance is affected 
when more data are collected. Overall the best performance was obtained with the ANN-VP with MO when using the features 
selected with the CFS technique.

Tables \ref{tab:molambda} and \ref{tab:mulambda} report the performance of the ANN-VP with minority oversampling 
and the ANN-VP with majority undersampling respectively with different number of categories $\lambda$ for their 
taxonomy using the two feature subsets. We can see that the number of categories does not affect the results to a 
big degree. However the values above $6$ or $7$ seem to give a somewhat better performance than smaller ones. 


\subsection{On-line Setting Results}

\begin{figure}
	\centering
		\subfloat[5 Hidden Units with $\lambda = 2$]{\includegraphics[trim = 4mm 4mm 4mm 8mm, clip, width=6.5cm]{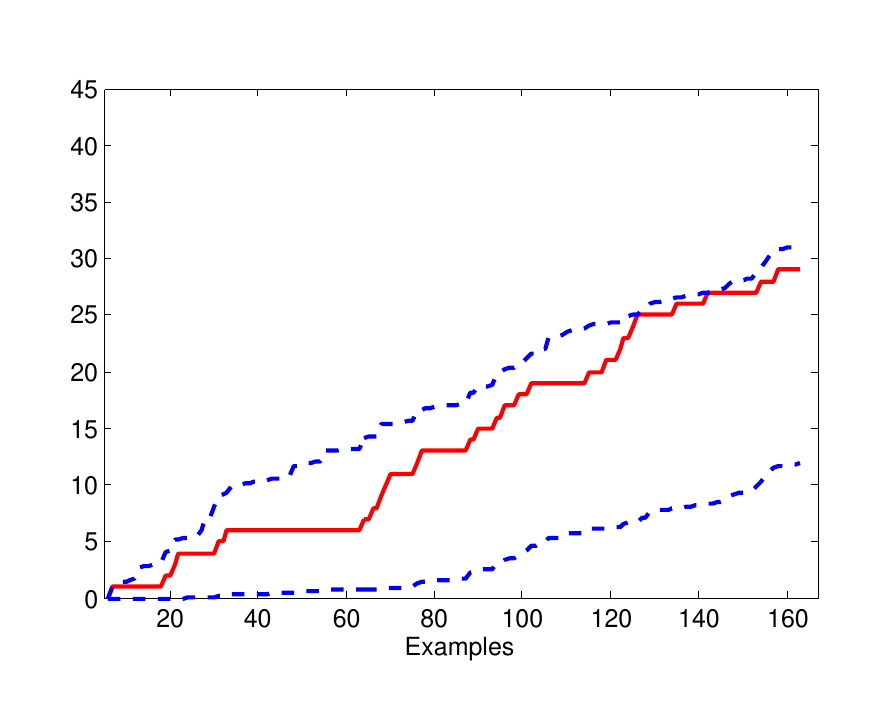}}
		\subfloat[100 Hidden Units with $\lambda = 2$]{\includegraphics[trim = 4mm 4mm 4mm 8mm, clip, width=6.5cm]{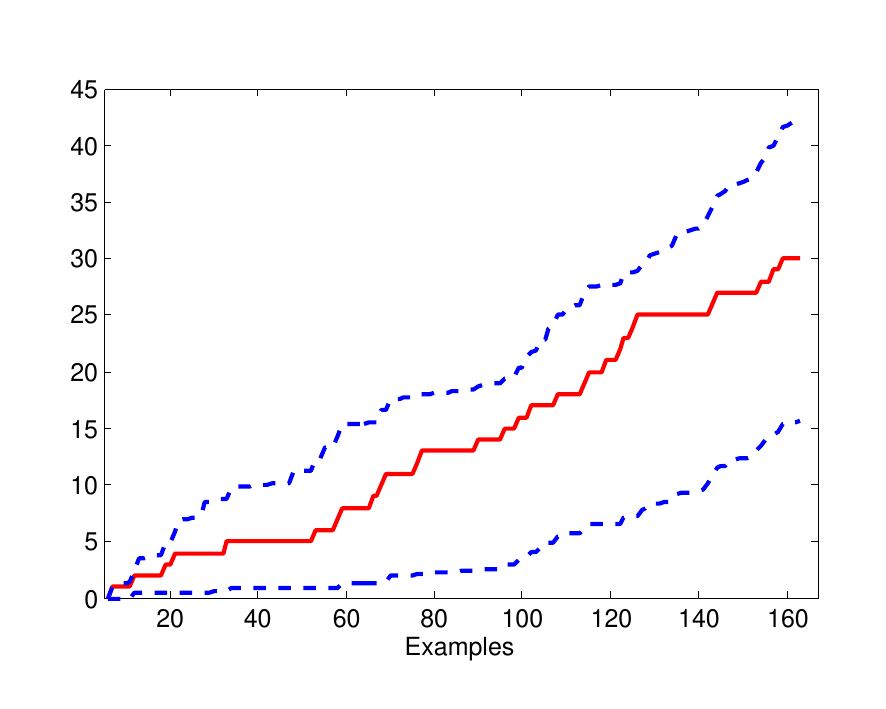}}\\
		\subfloat[5 Hidden Units with $\lambda = 6$]{\includegraphics[trim = 4mm 4mm 4mm 8mm, clip, width=6.5cm]{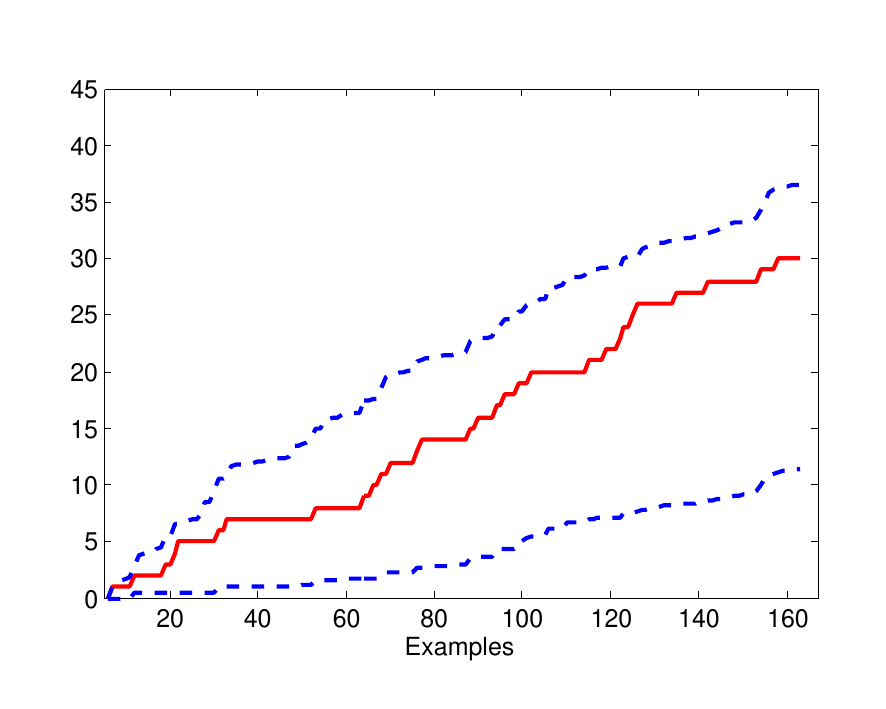}}
		\subfloat[100 Hidden Units with $\lambda = 6$]{\includegraphics[trim = 4mm 4mm 4mm 8mm, clip, width=6.5cm]{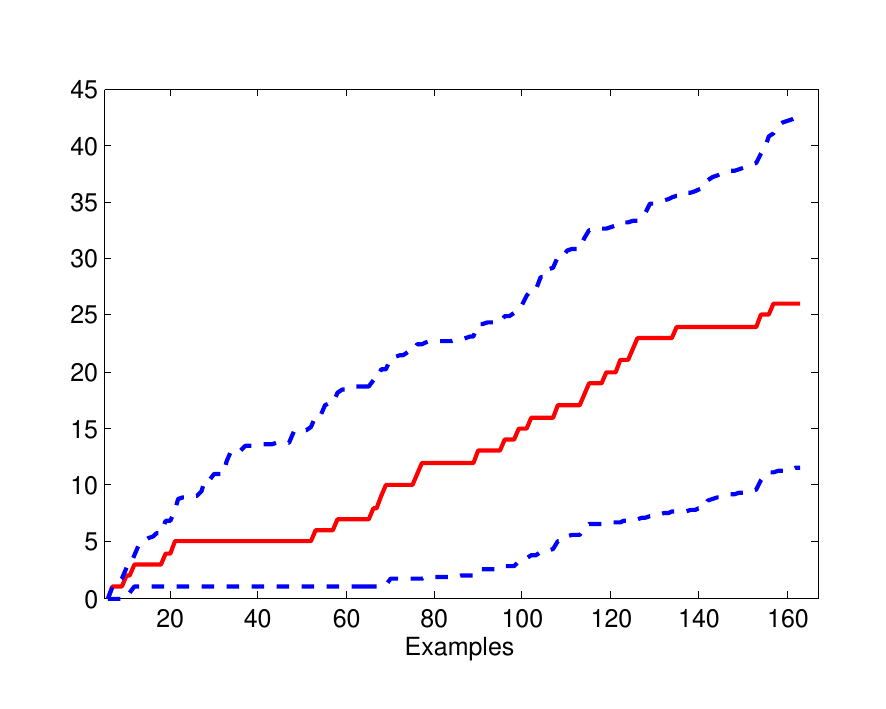}}\\
		\subfloat[5 Hidden Units with $\lambda = 10$]{\includegraphics[trim = 4mm 4mm 4mm 8mm, clip, width=6.5cm]{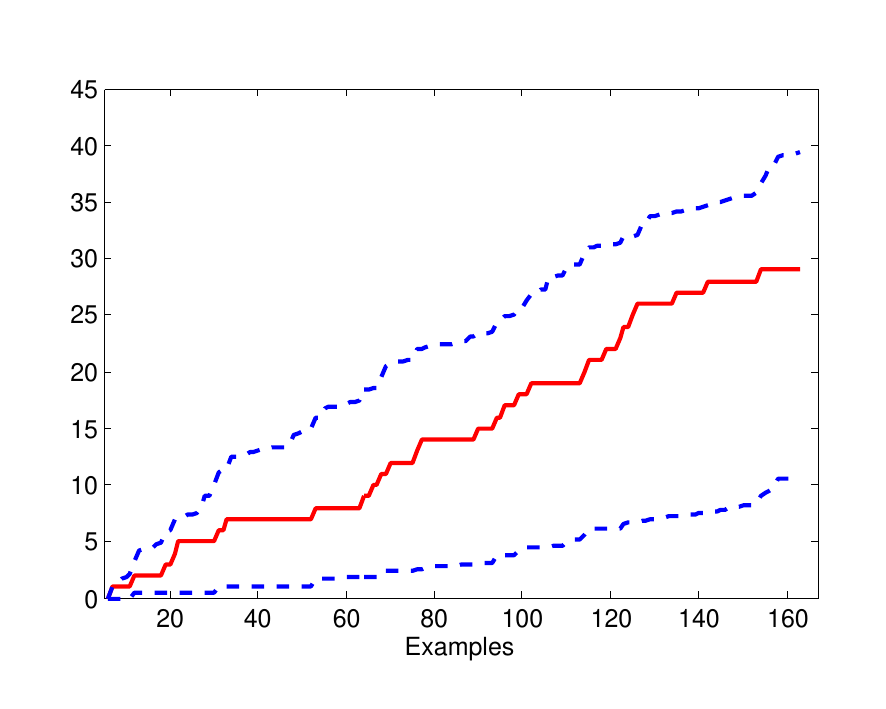}}
		\subfloat[100 Hidden Units with $\lambda = 10$]{\includegraphics[trim = 4mm 4mm 4mm 8mm, clip, width=6.5cm]{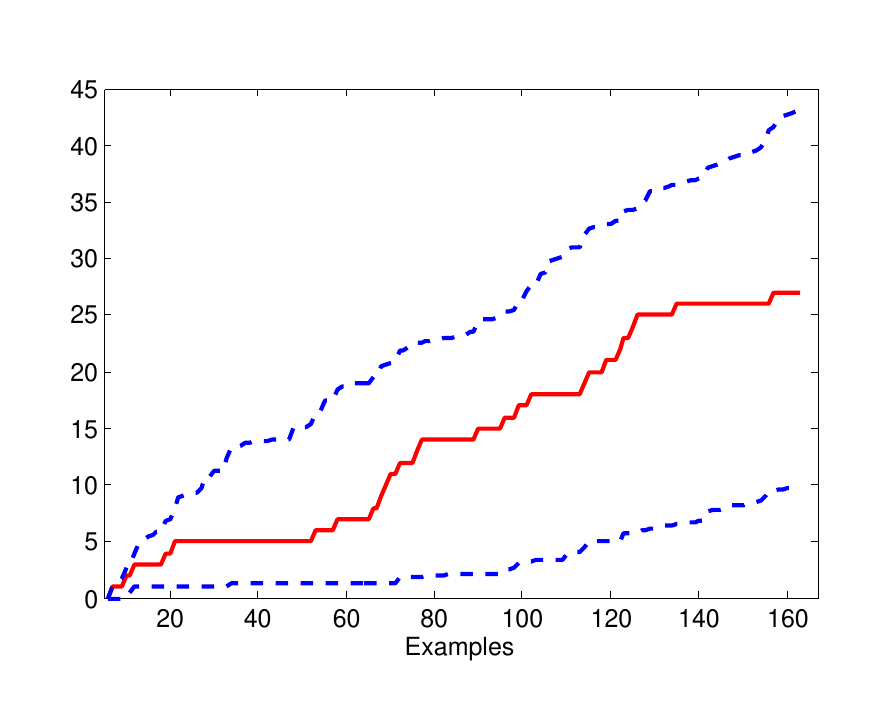}}		
\caption{On-line performance of ANN-VP with MO using the feature subset selected by CFS. Each plot shows the 
         cumulative number of errors $E_n$ with a solid line and the cumulative lower and upper error probability curves
         $LEP_n$ and $UEP_n$ with dashed lines.}
\label{fig:onlineVenn5at}
\end{figure}

\begin{figure}
	\centering
		\subfloat[5 Hidden Units with $\lambda = 2$]{\includegraphics[trim = 4mm 4mm 4mm 8mm, clip, width=6.5cm]{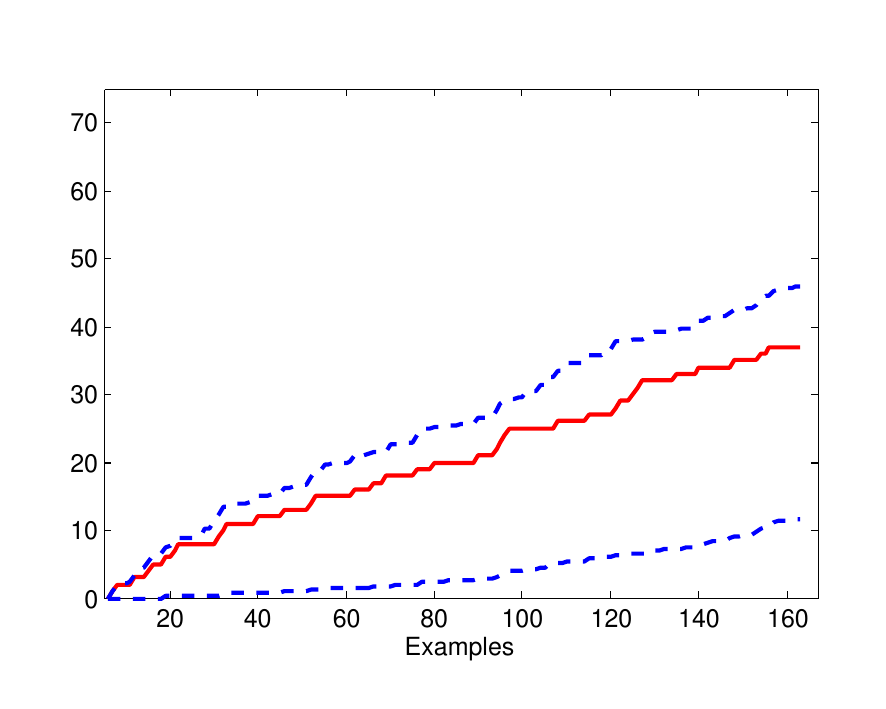}}
		\subfloat[100 Hidden Units with $\lambda = 2$]{\includegraphics[trim = 4mm 4mm 4mm 8mm, clip, width=6.5cm]{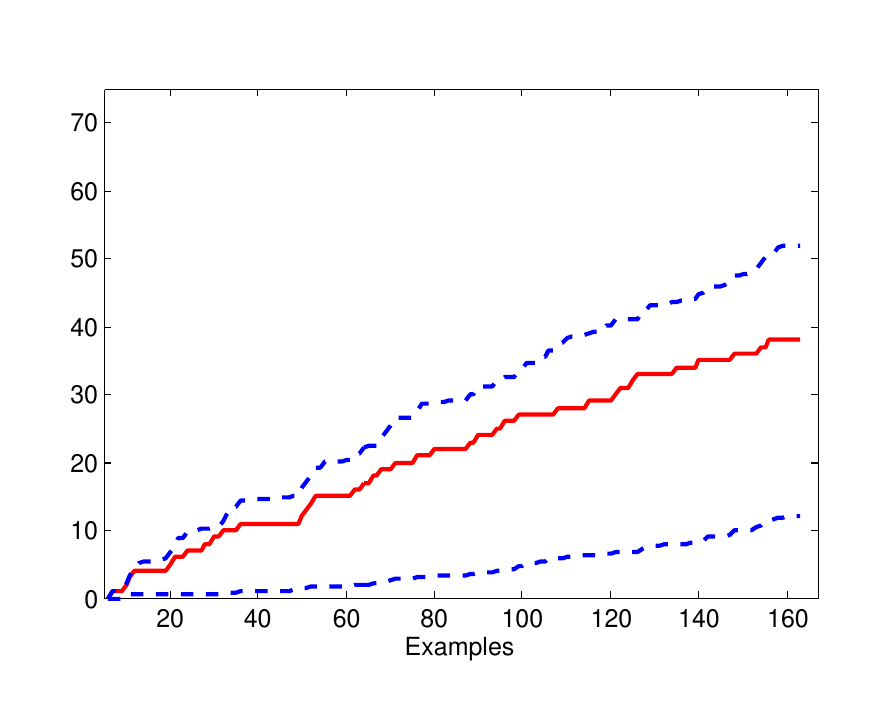}}\\
		\subfloat[5 Hidden Units with $\lambda = 6$]{\includegraphics[trim = 4mm 4mm 4mm 8mm, clip, width=6.5cm]{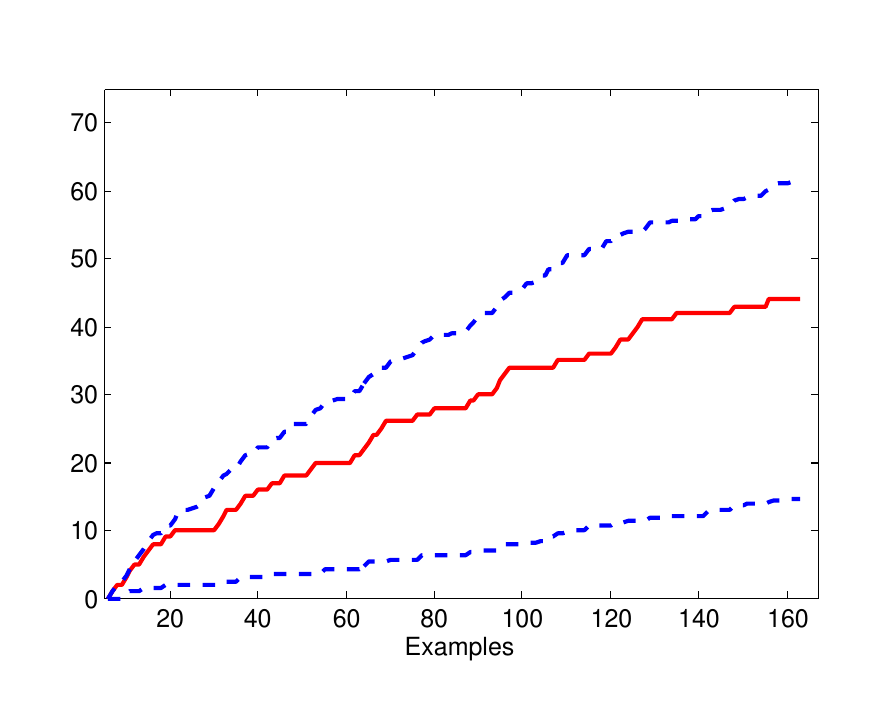}}
		\subfloat[100 Hidden Units with $\lambda = 6$]{\includegraphics[trim = 4mm 4mm 4mm 8mm, clip, width=6.5cm]{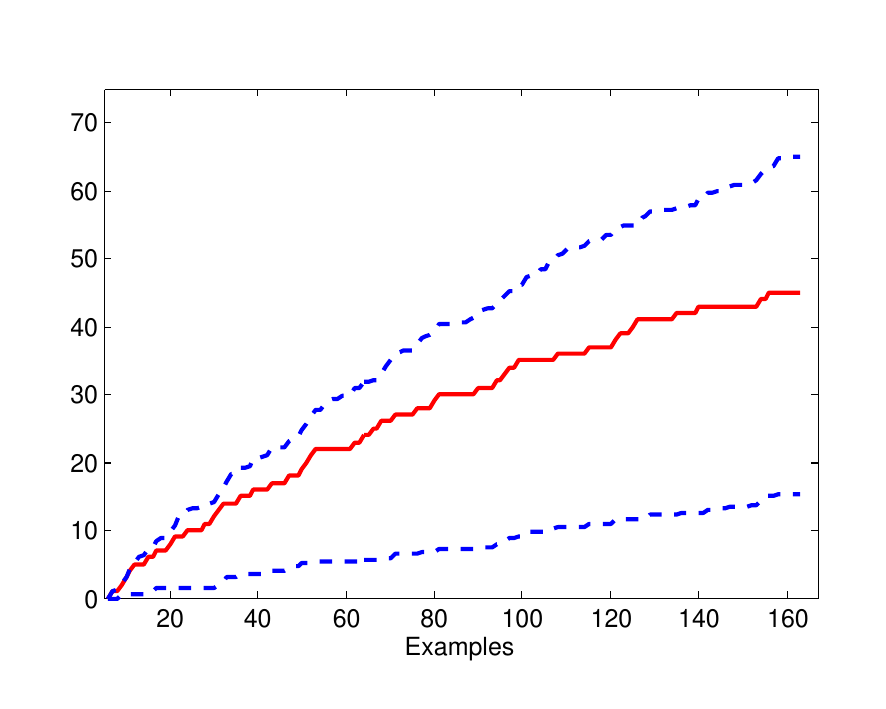}}\\
		\subfloat[5 Hidden Units with $\lambda = 10$]{\includegraphics[trim = 4mm 4mm 4mm 8mm, clip, width=6.5cm]{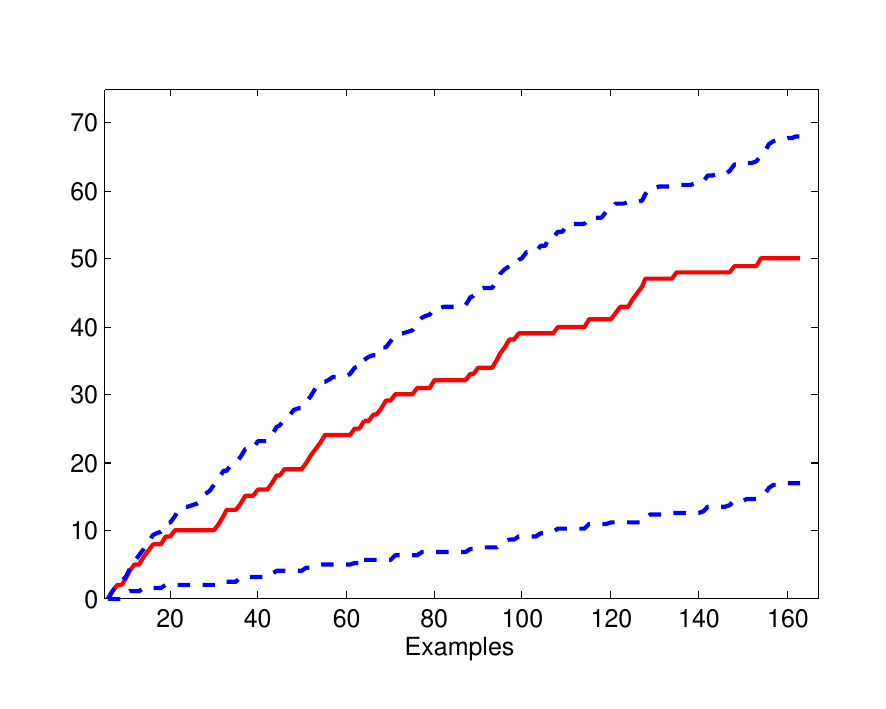}}
		\subfloat[100 Hidden Units with $\lambda = 10$]{\includegraphics[trim = 4mm 4mm 4mm 8mm, clip, width=6.5cm]{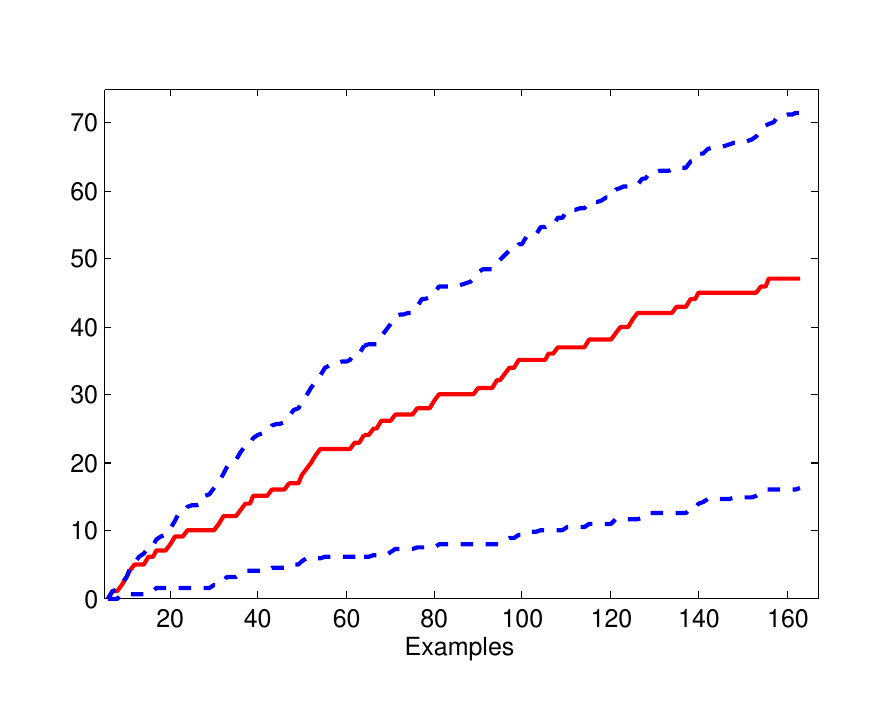}}		
\caption{On-line performance of ANN-VP with MO using the feature subset selected by $\chi^2$ and IG. Each plot shows the 
         cumulative number of errors $E_n$ with a solid line and the cumulative lower and upper error 
         probability curves $LEP_n$ and $UEP_n$ with dashed lines.}
\label{fig:onlineVennIG}
\end{figure}

\begin{figure}[t]
	\centering
		\subfloat[10 Hidden Units - CFS features]{\includegraphics[trim = 4mm 4mm 4mm 8mm, clip, width=6.5cm]{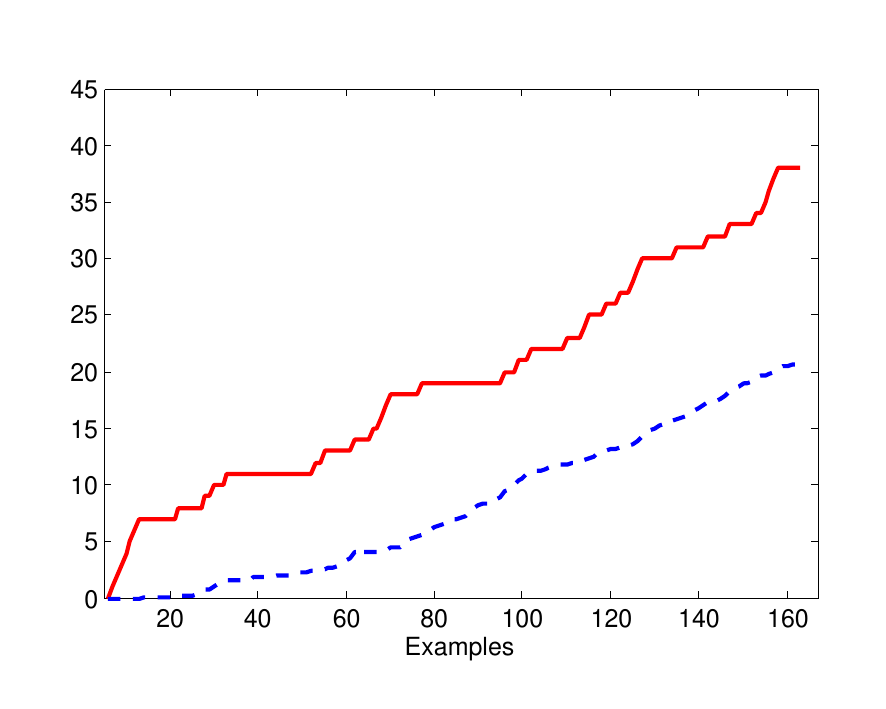}}
		\subfloat[100 Hidden Units - CFS features]{\includegraphics[trim = 4mm 4mm 4mm 8mm, clip, 
width=6.5cm]{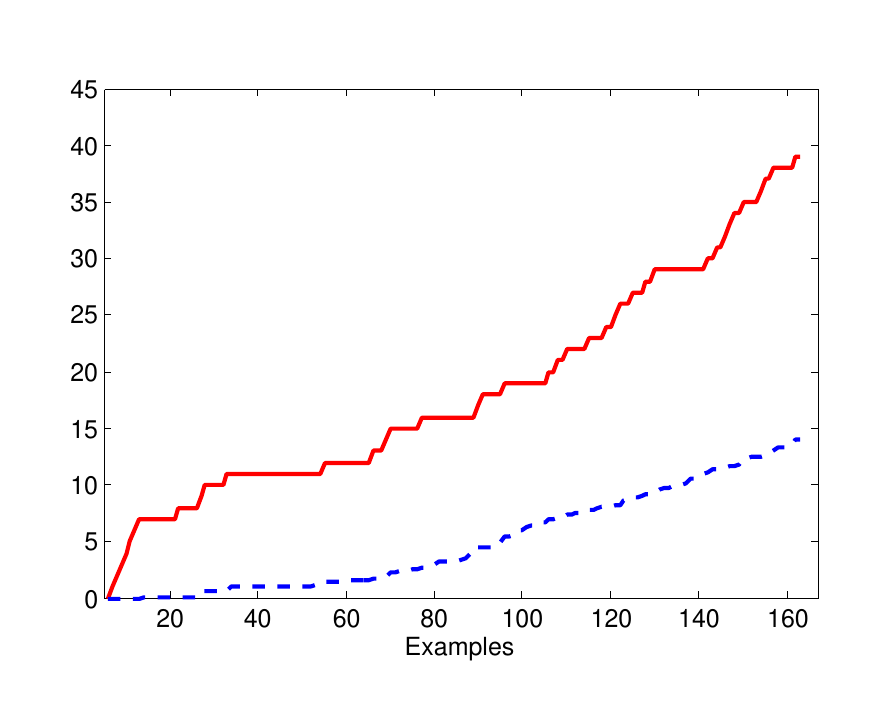}}\\
		\subfloat[10 Hidden Units - $\chi^2$/IG features]{\includegraphics[trim = 4mm 4mm 4mm 8mm, clip, width=6.5cm]{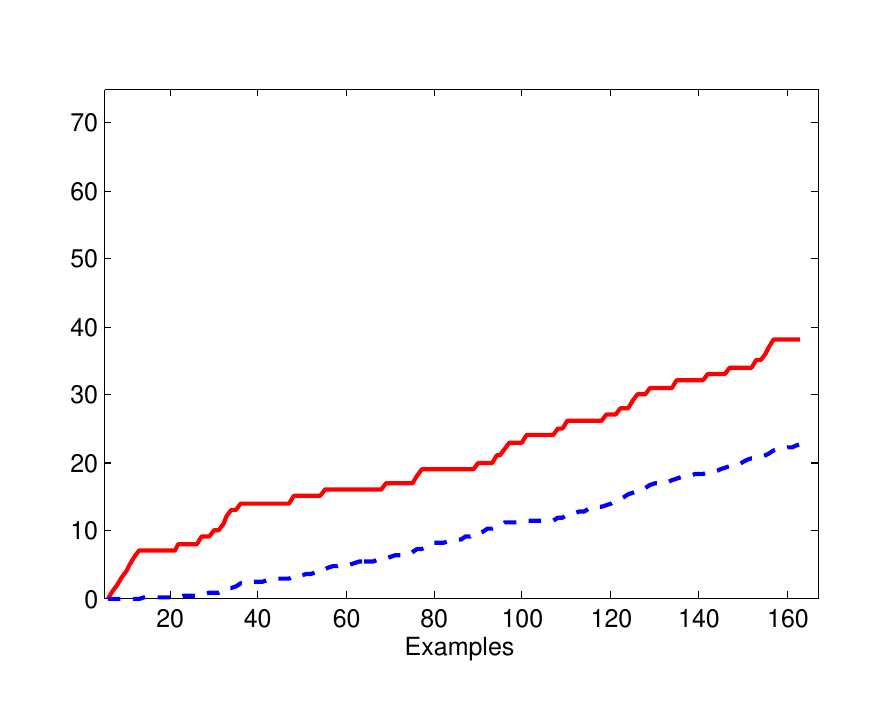}}
		\subfloat[100 Hidden Units - $\chi^2$/IG features]{\includegraphics[trim = 4mm 4mm 4mm 8mm, clip, 
width=6.5cm]{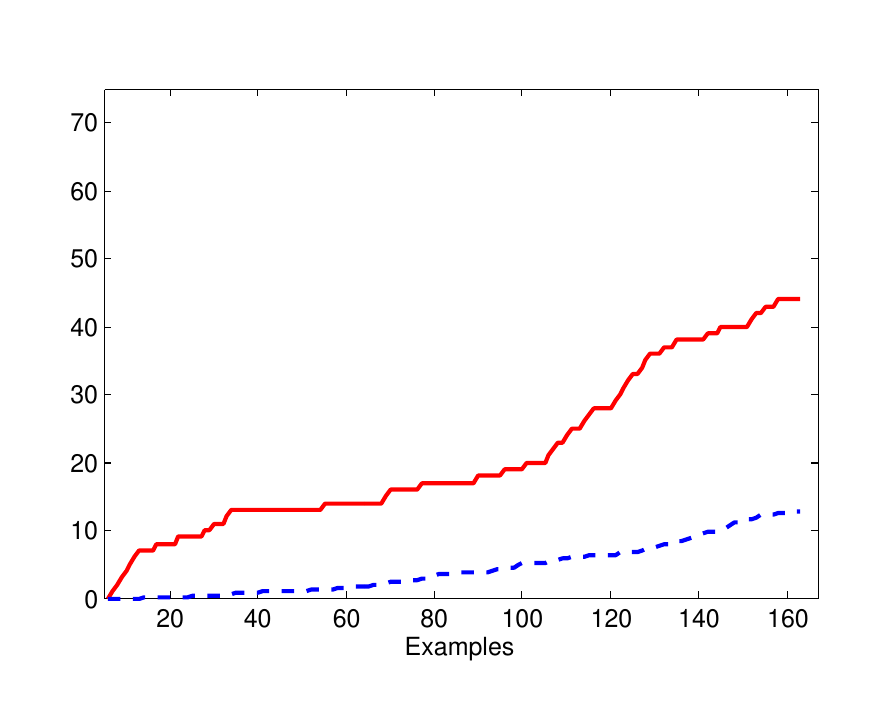}}\\
\caption{On-line performance of the original ANN with MO classifier. Each plot shows the 
         cumulative number of errors $E_n$ with a solid line and the cumulative error probability curve 
         $EP_n$ with a dashed line.}
\label{fig:onlineNN}
\end{figure}

This subsection presents the results obtained when applying the ANN-VP and conventional ANN 
approaches with minority oversampling in the on-line setting; we examine the minority oversampling approach 
as it gave the best results in the batch setting. Specifically, in this setting each experiment started with 
an initial training set of $5$ 
examples and one by one the remaining $157$ examples were predicted in turn and immediately after prediction 
their true classification was revealed and they were added to the training set for predicting the next example. 
In order to demonstrate that the choice of hidden units and number of taxonomy categories $\lambda$ does not 
affect the validity of the resulting probabilistic outputs, we performed this experiment with $5$ and $100$ hidden 
units and with $\lambda$ set to $2$, $6$ and $10$.
The results are presented in Figures~\ref{fig:onlineVenn5at} and \ref{fig:onlineVennIG} in the form of the 
following three curves for each experiment:
\begin{itemize}
\item the cumulative error curve
\begin{equation}
\label{eq:cumErr}
    E_n = \sum^n_{i=1} err_i,
\end{equation}
where $err_i = 1$ if the prediction $\hat y_i$ is wrong and $err_i = 0$ otherwise,
\item the cumulative lower error probability curve
\begin{equation}
\label{eq:cumLEP}
LEP_n = \sum^n_{i=1} 1 - U(\hat y_i)
\end{equation}
\item and the cumulative upper error probability curve
\begin{equation}
\label{eq:cumUEP}
UEP_n = \sum^n_{i=1} 1 - L(\hat y_i).
\end{equation}
\end{itemize}
The plots on the left present the results with $5$ hidden units while the plots on the right present the results
with $100$ hidden units. In terms of the number of taxonomy categories the top plots present the results with $2$ 
categories, the middle plots present the results with $6$ categories and the bottom plots present the results with 
$10$ categories. The plots confirm that the probability intervals produced by ANN-VP are well-calibrated since the 
cumulative errors are always included inside the cumulative upper and lower error probability curves. 
This shows that the ANN-VP will produce well-calibrated upper and lower error probability bounds regardless of the 
choice of features and taxonomy parameters. Note that although the use of the feature subset selected by 
$\chi^2$ and IG (figure~\ref{fig:onlineVennIG}) gives a higher number of errors than the one generated with 
CFS (figure~\ref{fig:onlineVenn5at}), the bounds of the VP remain well-calibrated, they just become 
wider to accommodate the higher uncertainty. Also the bounds generated with $100$ hidden units are again wider than 
those generated with $5$. Finally it seems that increasing the number of taxonomy categories also gives wider 
probability bounds possibly due to the relatively small size of the dataset.

The same experiment was performed with the original ANN classifier (with MO) on the two feature subsets and analogous 
plots are displayed in Figure~\ref{fig:onlineNN}. The two top plots present the results when using the feature subset 
selected by CFS while the two bottom plots present the results when using the feature subset selected 
by $\chi^2$ and IG. In this case the cumulative error curve (\ref{eq:cumErr}) is 
plotted together with the cumulative error probability curve
\begin{equation}
\label{eq:cumEP}
EP_n = \sum^n_{i=1} |\hat y_i - \hat p_i|,
\end{equation}
where $\hat y_i \in \{0, 1\}$ is the ANN prediction for example $i$ and $\hat p_i$ 
is the probability given by ANN for example $i$ belonging to class $1$. In effect 
this curve is a sum of the probabilities of the less likely class for each 
example according to the ANN. One would expect that this curve would be near 
the cumulative error curve if the probabilities produced by the ANN were well-calibrated. 
The plots of Figure~\ref{fig:onlineNN} show that this is not the case. The ANNs 
underestimate the true error probability in both cases since the cumulative error curve 
is much higher than the cumulative error probability curve. To check how misleading the probabilities 
produced by the ANN are, the $2$-sided $p$-value of obtaining the resulting total number of errors 
$E_N$ with the observed deviation from the expected errors $EP_N$ given the probabilities produced 
by the ANN was calculated for each case. The resulting $p$-values for the ANNs with $5$ hidden units 
were $0.000019458$ with the CFS feature subset and $0.0001704$ with the $\chi^2$/IG feature subset. 
The corresponding $p$-values for the ANNs with $100$ hidden units were even smaller; actually much smaller. 
This shows how misleading the probabilities produced by 
conventional ANNs can be, as opposed to the well-calibrated bounds produced by VPs.

\section{Conclusions}\label{sec:Conc}

This study applied Venn Prediction coupled with ANNs to the problem of VUR detection. 
In order to address the class imbalance problem of the particular task, minority oversampling 
and majority undersampling were incorporated in the ANN Venn Predictor. Unlike conventional 
classifiers the proposed approach produces lower and upper bounds for the conditional probability of 
each child having VUR, which are valid under the general i.i.d. assumption.

Our experimental results show the superiority of the VP approaches over conventional ANNs. The difference is 
especially significant in the case of the reliability metric, which points out that conventional ANNs can be 
quite unreliable as opposed to the proposed approaches. The proposed ANN-VP with MO outperforms all other methods 
while its comparison with the original ANN-VP approach shows that majority oversampling results in a big improvement
especially in terms of sensitivity.

Moreover our experimental results in the on-line setting demonstrate that the probability 
bounds produced by ANN-VP with MO are well-calibrated regardless of the feature subset used, 
number of taxonomy categories and number of hidden units used in the underlying ANN. 
On the contrary, the 
single probabilities produced by conventional ANN were shown to be very misleading.

Based on these results we believe that the proposed approach can be used in clinical practice for supporting the 
decision of whether a child should undergo further testing with VCUG or not. The use of this approach does not require 
any specialized tests other than the ones already being performed, therefore it can be easily incorporated into 
clinical practice without any significant costs. But most importantly the guaranteed reliability of the 
probabilistic bounds it produces means that they can be used by clinicians for taking informed decisions without the 
risk of being misled as cases with high uncertainty will be indicated either by very wide probabilistic bounds or 
by probabilities near 0.5. Therefore in cases with probabilistic bounds near 0 (negative for VUR) the decision that 
a VCUG is not needed can be taken with rather high certainty.

The main directions for future work include the assessment of the proposed approach in clinical practice and the collection 
of a bigger dataset, which will allow drawing more definite conclusions. Finally, experimentation with VPs based on other 
conventional classifiers is also in our future plans.

\end{document}